\documentclass{article}
\usepackage{arxiv}
\usepackage[utf8]{inputenc}
\usepackage{amsmath,amssymb,amsfonts}
\usepackage{algorithmic}
\usepackage{graphicx}
\usepackage{textcomp}
\usepackage{xcolor}
\usepackage{booktabs}
\usepackage{makecell}
\usepackage{balance}
\usepackage[nolist]{acronym}
\usepackage{bm}
\usepackage{moresize}
\usepackage{url}
\usepackage{tikz}
\usepackage[most]{tcolorbox}
\usepackage{graphicx}
\usepackage{subfig}
\usepackage{url}
\usepackage{enumitem}
\usepackage{cuted}
\usepackage{caption}
\usepackage{xcolor,colortbl}
\definecolor{Gray}{gray}{0.85}
\newcolumntype{C}{>{\centering\let\newline\\\arraybackslash\hspace{0pt}}m{0.1\textwidth}}
\newcolumntype{A}{>{\centering\let\newline\\\arraybackslash\hspace{0pt}}m{0.05\textwidth}}
\newcolumntype{V}{>{\let\newline\\\arraybackslash\hspace{0pt}}m{0.38\textwidth}}
\newcolumntype{L}{>{\let\newline\\\arraybackslash\hspace{0pt}}m{0.57\textwidth}}
\newcolumntype{B}{>{\let\newline\\\arraybackslash\hspace{0pt}}m{0.82\textwidth}}
\newcolumntype{M}{>{\let\newline\\\arraybackslash\hspace{0pt}}m{0.1\textwidth}}

\begin{document}

\newcommand{\trtitle}{Affective Computing for Human-Robot Interaction Research: Four Critical Lessons for the Hitchhiker}
\newcommand{\nl}{\vspace{0.5mm}{\noindent}}

\title{\trtitle}

\author{
Hatice Gunes and Nikhil~Churamani\\
Department of Computer Science and Technology\\ University of Cambridge, United Kingdom.\\
\texttt{Email: hatice.gunes@cl.cam.ac.uk}
}

\maketitle

\begin{abstract}
Social Robotics and \acf{HRI} research relies on different \acf{AC} solutions for sensing, perceiving and understanding human affective behaviour during interactions. This may include utilising \textit{off-the-shelf} affect perception models that are pre-trained on popular affect recognition benchmarks and directly applied to situated interactions. However, the conditions in situated human-robot interactions differ significantly from the training data and settings of these models. Thus, there is a need to deepen our understanding of how \ac{AC} solutions can be best leveraged, \textit{customised} and applied for situated \ac{HRI}. This paper, while critiquing the existing practices, presents \textit{four critical lessons} to be noted by the hitchhiker when applying \ac{AC} for \ac{HRI} research. These lessons conclude that:~(i)~The six basic emotions categories are irrelevant in situated interactions,~(ii)~Affect recognition accuracy (\%) improvements are unimportant,~(iii)~Affect recognition does not generalise across contexts, and~(iv)~Affect recognition alone is insufficient for adaptation and \textit{personalisation}. By describing the background and the context for each lesson, and demonstrating how these lessons have been learnt, this paper aims to enable the hitchhiker to successfully and insightfully leverage \ac{AC} solutions for advancing \ac{HRI} research. 

\end{abstract}

\section{Introduction \& Background}
\label{sec:intro}

Social Robotics has emerged as an inherently multi-disciplinary field bringing together research efforts from \acf{AC}, Social Signal Processing (SSP), \acf{CV}, \acf{ML} and \acf{HRI}. Yet, there is a need to develop affect sensing, perception and understanding methodologies targeted specifically to facilitate social robotics applications. To avoid re-inventing the wheel, researchers within the \acf{HCI}, \ac{HRI} and Social Robotics fields often, and rightly so, utilise available \textit{off-the-shelf} sensing or perception tools from other domains (such as face and gesture recognition) directly for their in-house studies, datasets and evaluations. However, these practices hinder progress leading to a lack of novel and domain-specific (affect) sensing, learning and adaptation algorithms. Furthermore, it impedes measures for reproducibility~\cite{GunesEtAl-Repro2022} due to a lack of purposeful, naturalistic and publicly available (affect) models, datasets and metrics, which are vital for comparative evaluation and gathering insights to push the field forward towards real-world adoption.

Recent research discussions\footnote{Discussions following a keynote address at the 3rd Workshop on Applied Multimodal Affect Recognition (AMAR), \acf{ICPR} 2022.} around situated affective computing, have emphasised understanding the role of \ac{AC} research, especially in situated interactions, and in realising social and affective interactions with robots. It is essential to appreciate what \textit{does not work} when undertaking situated \ac{AC} research and what \textit{lessons} we can learn from these failures. Furthermore, linking these lessons to \ac{HRI} research\footnote{Discussions following workshop keynote addresses at IEEE Int'l Conference on Robot \& Human Interactive Communication (RO-MAN'22), and the AAAI Fall Symposium on Artificial Intelligence for Human-Robot Interaction 2022.}, it is critical to understand how advances in (affect) sensing, perception and understanding mechanisms influence how individuals interact with social robots. The aim of this paper, thus, is to provide the hitchhiker with a guide for leveraging \ac{AC} for \ac{HRI} research based on the critical lessons learnt, both from \textit{successes} and \textit{failures}, grounded in and distilled from a broader set of \ac{AC} research studies conducted under situated interaction settings. Such a guide aims to inform the \ac{HRI} community, especially the hitchhikers starting their \ac{HRI} research journey, \textit{what to be aware of} when applying \ac{AC} tools for \ac{HRI} research. Similar recommendations have been compiled and shared as advice to aspiring experimenters on child-robot interaction in the wild~\cite{RosEtAl-ICMI2011}.

\begin{figure*}[t!]
	\centering
	\includegraphics[width=\textwidth]{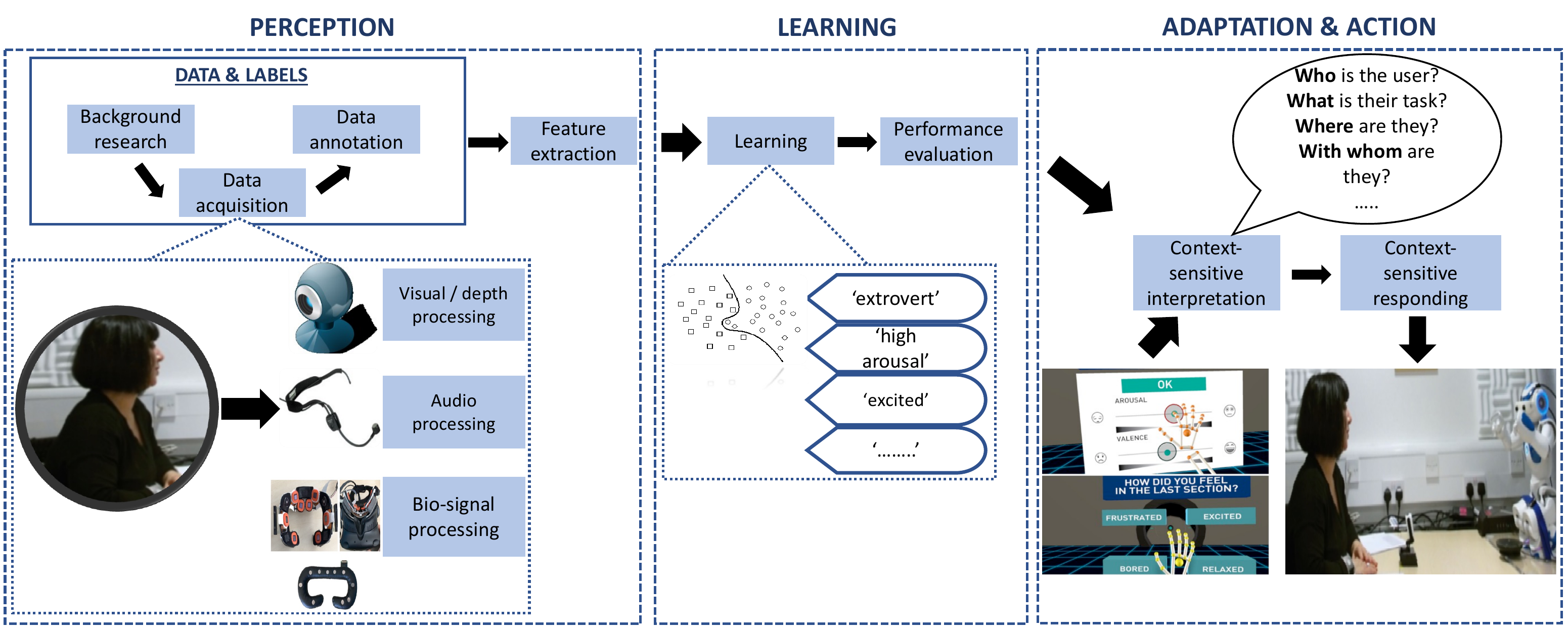}
	\caption{The pipeline for Affective Computing for Human-Robot Interaction implementations with marked stages of \textbf{Perception}, \textbf{Learning}, and \textbf{Adaptation \& Action}.}
	\label{ACPipeline}
    \vspace{-3mm}
\end{figure*}
This paper discusses four \textit{critical lessons} learnt applying \ac{AC} tools for \ac{HRI} research, especially for situated interactions. For each lesson, along with the \textit{background} understanding, a detailed account is provided of the \textit{context} under which the lesson is learnt, with \textit{explanations and insights} gathered, linking it to an \ac{HRI} context. These lessons are:

\begin{itemize}[leftmargin=1.65cm]
    \item [\textbf{Lesson 1:}] The six basic emotion~\cite{Ekman1971Constants} categories (\textit{happiness, sadness, surprise, fear, anger} and \textit{disgust})  are (mostly) irrelevant in situated interactions;
    \item [\textbf{Lesson 2}:] Affect recognition accuracy (\%) improvements are unimportant;
    \item [\textbf{Lesson 3}:] Affect recognition does not generalise (well) across contexts (e.g., user, task, etc. \-- using the definition of \emph{context} in ~\cite{VinciarelliPB09});
    \item [\textbf{Lesson 4}:] Affect recognition alone is insufficient for adaptation and personalisation.
\end{itemize}

The overall pipeline (with the different stages) of \ac{AC} for \ac{HRI} implementations is illustrated in Fig.~\ref{ACPipeline}. 
Lesson~1~(Section~\ref{sec:lesson1}) relates to how the user data acquired during the interactions are annotated or labelled, while Lesson~2~(Section~\ref{sec:lesson2}) relates to the robot's perception of the user. Lesson~3~(Section~\ref{sec:lesson3}) corresponds to robot learning and Lesson~4~(Section~\ref{sec:lesson4}) corresponds the robot's adaptation and actions. Section~\ref{sec:conclusion} summarises the contributions of this paper as well as reflects upon the need for a critical review of existing \ac{AC} solutions of affective \ac{HRI} studies. 

\section{Lesson 1: Six Basic Emotion Categories are Mostly Irrelevant}
\label{sec:lesson1}

\subsection{Background}
Within the pipeline of creating an automatic affect recogniser, this lesson relates to the aspect of \emph{affect annotations and labels} (see \textit{Data \& Labels} under Fig.~\ref{ACPipeline}: \textit{Perception}). When researchers purchase or acquire commercial social robots, these robots come with black-box perception capabilities, one of which is usually proudly claimed to enable `automatic emotion recognition' for the robot. For instance, one of the features listed for Pepper Robot is `recognising emotions on your face'\footnote{\url{https://www.gwsrobotics.com/why-pepper-robot}}. In such robotic platforms, this means the recognition of the six basic emotion categories, namely, (neutral+) happiness, sadness, surprise, fear, anger and disgust~\cite{Ekman1971Constants}. However, this works only partially and under \textit{posed} expression settings. Real-world interactions are much more complex resulting in the robot struggling to accurately capture individual expressions, as demonstrated via interactive public demonstrations~\cite{GunesEtAl-2019}. But even then, it is important to evaluate and understand what such categorization of user affective behaviour means for situated human-robot interactions.


\subsection{Context}
Much of \ac{AC} research employs \ac{ML}-based automatic affect recognition models trained and benchmarked on publicly available datasets, acquired outside of situated interaction settings. For example, most \ac{CV} models undertaking the task of \ac{FER} are trained on static images crawled from the internet, with cropped facial regions where the situational context information has been removed, with crowdsourced labels corresponding to the aforementioned six basic emotion categories
Thus, as soon as these models are embedded in robotic systems for realistic applications including tutoring and learning, assistance with rehabilitation or physical and mental health, these models cannot cope with the variation and noise in the input data that they have not encountered in their training. This results in \ac{FER} often failing in situated human-robot interactions.

Beyond automatic recognition, the six basic emotion categories are widely used in various \ac{HRI} studies, even when these labels do not seem relevant for real-world contexts. One recent example investigating emotion perception using the basic emotion categories is an \ac{HRI}-based rehabilitation scenario, as this is expected to improve the experience of the patients~\cite{ViolaFMKC22}. 
In this study, a robotic arm is used to investigate whether and how it can communicate an emotional state through movements and whether people can attribute these movements to the intended emotional state. It found that \textit{happiness} was identified well, but not \textit{sadness} and \textit{anger}. However, going beyond these findings, it is important to understand \textit{`What does it mean for a robotic arm to display anger?'}. Furthermore, it is also important to understand \textit{`How useful are basic emotion categories for rehabilitation robotics?'} and \textit{`What implications does this have for HRI, in general?'}.

\subsection{Lesson \& Insights}

A critical evaluation of the questions posed above requires a deeper and fundamental understanding of the situational and contextual attributes that determine human behaviour during interactions. One needs to go beyond the six basic emotion categories and start exploring other affect and emotion models and instruments, while also considering how to use these contemporary models throughout the entire pipeline of study design, data acquisition, data annotation, and training and evaluation of \ac{ML} models. In doing so, it is essential to start with fundamental questions, such as \textit{`Which emotion or affect model is best suited to represent human behaviour and how do we decide this?'} Additionally, it is also important to consider \textit{`Whether we are taking into account situational or contextual aspects?'}.

Two contemporary instruments that can be used, instead of the six basic categories of emotions, are the \acf{SAM}~\cite{BRADLEY199449,SAM} and the \acf{GEW}~\cite{GEW}. \ac{SAM} is a picture-based questionnaire to independently evaluate the affect dimensions of \textit{arousal} (activation), \textit{valence} (pleasure) and \textit{dominance} (sense of control), and it can be used for subjective assessment of participant/user affective responses~\cite{SAM}. The \acs{GEW}, on the other hand, has been proposed as `a theoretically derived and empirically tested instrument to measure emotional reactions to objects, events, and situations'~\cite{GEW}. The participant/user can indicate the emotion they experienced by choosing a single emotion with the corresponding intensity or a blend of multiple emotions (out of $20$ emotion families). Robotics and \acs{HRI} researchers have started to successfully use \ac{SAM} and \ac{GEW} in their works, for example, to evaluate patients' emotions induced by a robotic hand rehabilitation platform~\cite{CisnalEtAl-2022}, to classify the expression of emotion on robots~\cite{McGinn-HRI18} and to measure perceived affect in \ac{HRI}~\cite{CoyneEtAl-HRI20}.

In the context of dyadic human-human interactions \textit{vs.} human-agent interactions, Song~et~al.~\cite{SongEtAl-TAC2022} report that facial reaction prediction and personality recognition performance for \ac{ML} models are better for human-human interaction data. This finding indeed has implications for \acs{HRI} research and brings forth further questions that, as a community, we would need to investigate. These include, but are not limited to, \textit{`Do we display affect differently in \ac{HRI}?'}, and \textit{'Do we need different affect or emotion models for \ac{HRI} that capture both qualitative and quantitative aspects of human as well as robot behaviours?'}.
In order to answer these questions, a promising direction is to take a data-driven approach, similar to the pioneering study by Jam~et~al.~\cite{JamEtAl-HRI21} that aims at developing a data-driven categorical taxonomy of emotional expressions in real-world \ac{HRI}.

\section{Lesson 2: Affect Recognition Accuracy (\%) improvements are unimportant}
\label{sec:lesson2}
\subsection{Background}
Within the pipeline of creating an automatic affect recogniser, this lesson relates to the aspect of \emph{affect sensing} (see \textit{Performance Evaluation} under Fig.~\ref{ACPipeline}). The majority of the work towards automatic affect recognition focuses on achieving results that are considered `excellent' or `very good' in terms of the evaluation metric used. For many researchers `success' is then equivalent to either obtaining a recognition accuracy of~$>=75\%$ on a dataset that perhaps other researchers have not yet worked or published on, or improving the \acf{SOTA} recognition accuracy by $>=2-3\%$ on a benchmark that others have widely reported on to be able to claim that their method is `better' than the current \ac{SOTA} results. However, benchmark datasets, even the ones that claim to be obtained \textit{in-the-wild}, are usually stripped of context. Such datasets, for instance, contain static facial images or even videos of people without much interaction taking place. When we move away from recognising affect on such \textit{in-the-wild} but idealised benchmark datasets to actual interaction studies with humans, we are faced with a much higher level of complexity. 

\begin{figure}[t!]
	\centering
	\includegraphics[width=0.5\textwidth]{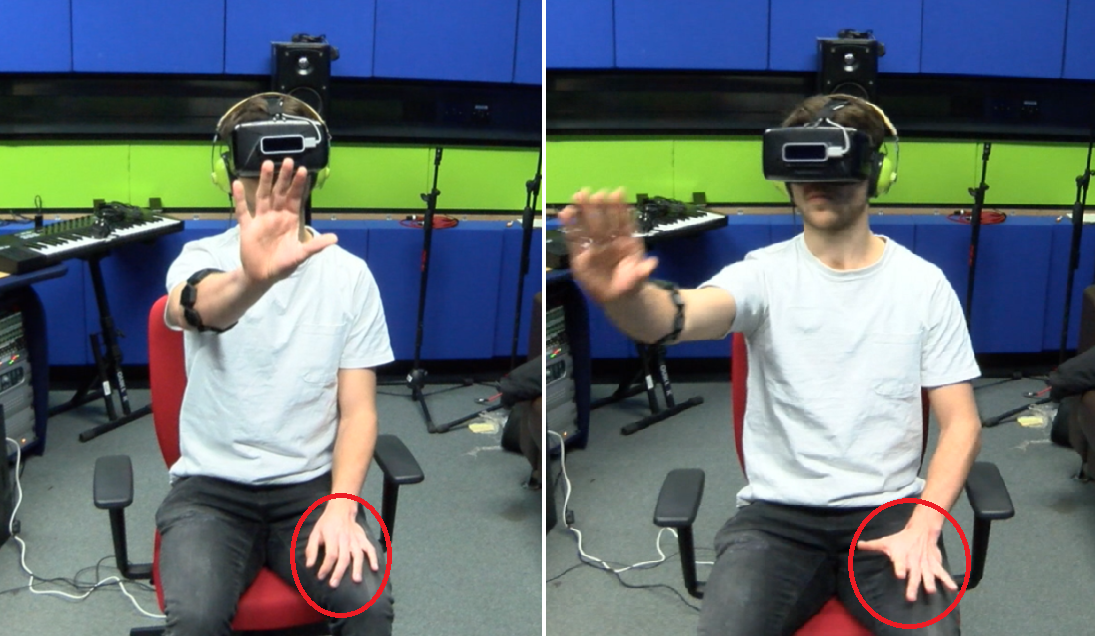}
	\caption{Illustrating the differences in participants' \textit{left hand} when playing the `Memory Break' game in \ac{VR} at \emph{Level 1~-~Easy} (left) \textit{vs.} \emph{Level 3~-~Hard} (right).}
	\label{LeftHand}
    \vspace{-3mm}
\end{figure}

\subsection{Context}

To exemplify how relying on affect recognition accuracy~(\%) improvements may be unimportant and insufficient in \ac{HRI} context, we look at \textit{`Gamified Cognitive Training'}, as an example, as it relates to one of our study~\cite{GabanaEtAl-ACII2017} undertaken in $2016-2017$. This study investigated how the affect dimensions of arousal and valence were linked to \acf{WM} performance of $30$ participants when playing a custom video game, `Memory Break', on Desktop \textit{vs.} in \acf{VR}, in two separate sessions, one for each interaction mode.
Both game modes were designed to have three difficulty levels to evoke different levels of \textit{arousal} while maintaining the same memory load. The \ac{WM} capacity baseline of participants were measured using relevant measures while the participants self-reported their affective states and completed the \acf{GEQ}~\cite{ijsselsteijn2013game}. Our analyses showed an improvement in participants' \ac{WM} performance when playing in \ac{VR} mode, with a significant effect in those with a low \ac{WM} capacity. Significantly higher levels of \textit{valence} and \textit{arousal} were self-reported when playing the \ac{VR} version of the game. 

To sense the participants' affective states, a heart-rate sensor was attached to their chest recording their heart activity and an \ac{EMG} armband was placed on the forearm that was used for interacting with the game environment. 
However, we had missed one important factor. As seen in Fig.~\ref{LeftHand}, when the difficulty level of the game increased to `hard' (Level 3), the tension was clearly observable on their hand that was resting on their lap. Post-study, we observed this to be the trend for all participants. Unfortunately, that hand did not have any sensor placed on it to measure the tension manifested, which meant we missed crucial information that could aid the recognition of participants' arousal and valence. Despite extracting features from other sensors and experimenting with various \ac{ML} techniques for classifying arousal and valence, the recognition results did not look promising. Ultimately, accuracy (\%) improvement in this context was unimportant because we were not measuring and analysing the most relevant signals.


\subsection{Lesson \& Insights}
The lesson that can be learnt from the `Gamified Cognitive Training' study (and other relevant ones) is that \emph{accuracy} or \emph{accuracy improvements (\%) are unimportant}, especially when we are not capturing and analysing the most relevant signals and cues. Expressly, undertaking human studies in situated interactions, where we aim to sense affect, requires several pilot study iterations, until we are sure about where to place the different sensors, measuring the right signals and cues related to the affect displayed. In other words, we should \emph{study the expressions and display of affect before we sense them}. This is mainly due to two reasons. Firstly, in the human-machine interaction context, humans shape machine behaviour and vice-versa~\cite{RahwanEtAlNature-2019}. This often results in the emergence of new human behaviour, unforeseen in the original study or interface design. Secondly, when analysing affect and emotions, there is, at times, non-verbal and emotional \textit{leakage}.  At times, inner feelings of a person might be revealed or expressed more intensely in a certain modality or channel~\cite{EkmanEtAl-Leakage-1969, WaxerEtAl-1977} (usually the less dominant one), which might be different from the one observers focus on, for example, controlling what is being said while expressing differently through nonverbal behaviour. In light of these, in situated interactions where we aim to measure and analyse socio-emotional behaviours, we need to reflect on critical questions such as \textit{`Are we placing the sensors in the right places?'}, and \textit{'Are we measuring the most relevant signals?'}.

With these aspects in mind, a possible direction for \ac{HRI} research can be to adopt rich multi-modal sensing, not necessarily to improve accuracy, but also to ensure that  different aspects of user affect and behaviour manifestations are captured and investigated. This is particularly important for emerging research areas that cannot simply rely on previous research findings. For instance, mental wellbeing evaluation in children via child-robot interactions~\cite{AbbasiSAFJG22} requires an investigation of different aspects of child multi-modal behaviour (questionnaire responses, free-from speech content, nonverbal head, face or audio behaviours and physiological reactions), going beyond what children report or say. However, it is important to note that not many social robots are equipped with high-resolution sensing or have the capabilities to enable such rich multi-modal perception. A possible solution may be to create `hacks', for example, by 3D printing and additional sensor placement (see~\cite{BremnerCG17} on how a 3D printed headset is used with high-resolution cameras).
%

\begin{figure}[t!]
	\centering
	\includegraphics[width=0.95\textwidth]{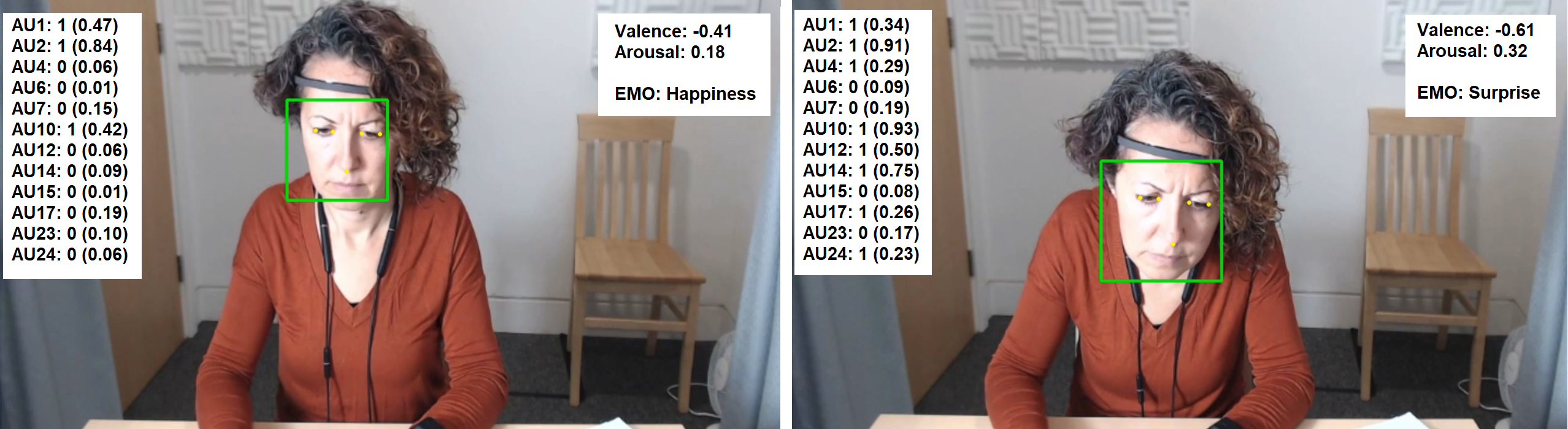}
	\caption{Predictions of (deep) \ac{ML} models trained on publicly available facial expression, facial affect and facial \acf{AU} datasets when used on data acquired under work-like settings and tasks.}
	\label{EU-WA-Face}
    \vspace{-3mm}
\end{figure}

\section{Lesson 3: Affect Recognition Does Not Generalise Well Across Contexts}
\label{sec:lesson3}

\subsection{Background}
Openly available facial affect datasets used for training \ac{FER} models generally contain displays of young and middle-aged adults. Facial affect data from sensitive user groups such as children, adolescents, and older adults are relatively less accessible due to various challenges including ethical and privacy concerns. This imbalance in data causes these models to not generalise well on other user groups such as the elderly~\cite{KaixinEtAl-2019} or children~\cite{HowardEtAl-2017} and, in turn, results in biased algorithms for facial affect analysis and prediction.
In addition to encoding demographic bias, currently available facial expression datasets are also biased towards certain affect labels such as ``neutral'', ``anger'' and ``happiness'', compared to other affective states such as ``annoyance''~\cite{CheongEtAl-2021}. 
Thus, models trained on most common benchmark datasets for facial affect recognition are: (i)~more accurate for young and middle-aged adults; and (ii)~mostly predicting affect in terms of basic emotion categories; despite the fact that this might not fit well the application context~\cite{CheongEtAl-2021}.

\subsection{Context}
To appreciate the challenges relating to applying \textit{generalised} affect recognition models, it is important to consider how these models may perform with under-represented (in traditional affect perception benchmarks) populations. In this context, the EU Horizon 2020 WorkingAge\footnote{\url{https://www.workingage.eu/}} project is aimed at studying and promoting healthy habits in working environments, focusing on people aged over $45$. By gathering a better understanding of wellbeing at work and of factors that may inhibit or deteriorate prolonged employment, it created an integrated digital solution, the \acf{WAOW} Tool, to support workers' wellbeing in three types of working environments: office, teleworking, and manufacturing. Within the \ac{WAOW} Tool pipeline of creating an automatic system that analyses worker psycho-social conditions, worker physical conditions and the working environment, and personalises via appropriate recommendations, customisable by the user~\cite{AlmeidaEtAl-Samos2020}, this lesson relates to the aspect of \emph{affect sensing} and \emph{recognition} (see Fig.~\ref{ACPipeline}: Perception).

As a part of the WorkingAge project, we first introduced a multi-site data collection protocol for acquiring human behavioural data under simulated working conditions with three work-like tasks: the N-back task, the video conference task and the operation game. With this, we acquired the first human working facial behaviour dataset called \emph{WorkingAge DB}~\cite{SongEtAl-2022} which was collected in four different sites across three countries in Europe (Germany, Italy, and the UK). 
Implementing (deep) \ac{ML} models (for example, ResNet-50), trained on publicly available facial expression (e.g., RAF-DB~\cite{li2019reliable}), facial affect (e.g., AffectNet~\cite{2018Affectnet}) and facial \ac{AU} (e.g., BP4D~\cite{ZHANG2014BP4D}) datasets, and applying these models on facial data acquired under work-like settings and tasks, results in evaluations similar to those illustrated in Fig.~\ref{EU-WA-Face}. It can be clearly seen that such models have no knowledge about context, and provide labels such as `surprise' and `negative valence' when the person is focused on the task.

Having seen these results, we decided to train \ac{ML} models specifically with the data acquired in work-like settings. Thus, in \cite{SongEtAl-2022}, we implemented and compared a set of (deep) \ac{ML} methods using the facial data from WorkingAge DB for automatic prediction of worker periodical facial affect while also investigating how task type, recording site, gender, and feature representations affected model performance. Our results showed that worker affect can be inferred from their facial behaviours using data acquired in work-like settings, and models pre-trained on naturalistic datasets are useful for prediction but are insufficient on their own. Context, specifically the task type and task setting, influenced the affect recognition performance~\cite{SongEtAl-2022}.


\subsection{Lesson \& Insights}

\ac{HCI} and \ac{HRI} studies are prone to adopting \textit{off-the-shelf} affect recognition toolkits, that are pre-trained on publicly available benchmark datasets, as means to an end, for the quick modelling of user affective behaviour. 
For instance, in the \ac{HRI} context, Mathur et al.~\cite{MathurSXLM21} investigated how to model user empathy elicited by a robot storyteller. For this, they employed an open-source off-the-shelf toolkit (OpenFace 2.2.0~\cite{OpenFace2018}) that is widely used by various researchers within the \ac{AC}, \ac{HRI} and \ac{HCI} communities. OpenFace enables the extraction of eye gaze directions, the presence (and intensity) of $17$ facial \acp{AU}, facial landmarks and head pose coordinates, amongst other features. 
However, as we learnt from the WorkingAge study~\cite{SongEtAl-2022}, for facial affect recognition in specific contexts such as work-like settings, we cannot simply rely on generic off-the-shelf toolkits. Such models are ignorant of context and will not generalise well to real-world settings where many factors (such as ethnic or cultural background, gender, age, and the task, amongst others) influence human expressivity and nonverbal behaviour. 

Thus, when analysing human affective behaviour using off-the-shelf toolkits and models, several critical questions need to be considered. These include, but are not limited to: \textit{`How well are we taking into account the contextual aspects of the interaction?'}, and \textit{`Are we considering person-specific aspects impacting the interacitons?'}. To address these questions, we need to focus on \textit{personalisation} rather than generalisation, considering person-specific aspects when modelling user affective behaviour. For example,~\cite{ParkGSGB19-AAAI} presents a personalised learning companion that uses children’s verbal and nonverbal affective cues to modulate their engagement levels. Facial features extracted using the off-the-shelf Affdex toolkit~\cite{Affdex2016} are used for arousal prediction which, in turn, defines state space features for an \ac{RL}-based personalisation algorithm. 
More recently, in~\cite{churamani2020continual} we introduced and adapted the \ac{CL} paradigm for Affective Robotics where a robot acquires and integrates knowledge incrementally about changing data conditions, and showed how it can be utilised in practice for adaptive \ac{HRI}~\cite{Churamani-ACIIW2022}. Furthermore, the series of \acs{LEAP-HRI}\footnote{https://leap-hri.github.io/} workshops, that we have been organizing since 2021, also emphasised the need to move away from generalisation and focus more on lifelong learning and personalisation, particularly when it comes to long-term \ac{HRI} where novelty effect is no longer present~\cite{IrfanRSPG-HRI22}.

Additionally, we also need to consider other relevant questions such as \textit{`Are we investigating for whom the trained models work well, and why?'}, \textit{`How do these models work for specific user groups like children and elderly?'}, and \textit{`What are we doing to ensure that predictions from these models are not biased?}'. 
To date, there are many publicly available benchmark datasets for expression and affect recognition, however, none of these datasets have been acquired considering a \textit{fair distribution} across the human population. Recent studies on a number of publicly available benchmark datasets such as RAF-DB~\cite{li2019reliable} and CelebA~\cite{liu2015faceattributes} have shown that \ac{ML} models for \ac{FER} trained on such datasets are biased~\cite{XuWKG-ECCVW20}. Despite several bias mitigation strategies~\cite{HowardEtAl-2017,XuWKG-ECCVW20,ChuramaniEtAl-TAC2022} for addressing biased model predictions in the \ac{AC} context (see~\cite{CheongEtAl-2021} for a review), how bias in affect prediction models impacts \ac{HRI} and user experience, engagement and trust, and how to achieve \emph{fairer affective robotics} remain open research problems that need multi-disciplinary community efforts at the level of datasets, annotations, benchmarking and reproducibility.

\section{Lesson 4: Affect recognition alone is insufficient for adaptation and personalisation}
\label{sec:lesson4}
\subsection{Background}
Within the pipeline for creating an automatic affect recogniser, this lesson relates to the aspect of \emph{adaptation} (see Fig.~\ref{ACPipeline}: Adaptation \& Action). 
Affect recognition is only one of the affective cognitive architecture modules for achieving emotionally intelligent autonomous robots that are capable of perception, learning, action, adaptation, and even anticipation~\cite{TanevskaEtAl-2018}. One of the most common techniques for robot learning and adaptation is learning with the human-in-the-loop, or \ac{IRL}, that focuses on sensing and incorporating user interactive (verbal, social and affective) feedback~\cite{leite2014empathic,GamborinoEtAl-ROMAN19,LiEtAl-2020,cui2020empathic,LinEtAl-2020, Gillet-HRI2022, McQuillinEtAl-HRI2022,Churamani2022AffCore}. \acs{IRL} with explicit feedback can be challenging as humans tend to provide more positive than negative feedback, at times ignoring the robots' mistakes. With progressing interactions, the frequency of human feedback may decrease~\cite{macglashan2017interactive}. Therefore, using implicit feedback, such as facial affect, can be more effective as the human ``teacher'' will be less conscious of providing the feedback and will be less likely to suffer from feedback fatigue~\cite{LinEtAl-2020}. Studies on \acs{IRL} demonstrate the growing potential of sensing and utilising implicit human behavioural cues, such as facial expressions and affect, for training robots through natural interactions and shaping their behaviour in real-time. But '\textit{is adaptation based on affect sufficient and does it always improve \ac{HRI} experience?}'

\subsection{Context}
For naturalistic \ac{HRI}, especially facilitating social interactions, it is imperative that robots are able to sense and adapt towards human behaviour, not only regarding individual responses as feedback on their actions but also as motivation for learning context-appropriate behaviours. In this context, in~\cite{McQuillinEtAl-HRI2022}, we explored learning socially appropriate Robo-waiter behaviours through real-time user feedback. This feedback was driven by either an implicit reward (calculated by observing the facial affective behaviour of participants) or an explicit reward (incorporating their verbal responses). First, a dataset was created and annotated using crowd-sourced labels to learn appropriate approach behaviours for a robo-waiter based on its positioning and movement. This dataset was then used to pre-train an \ac{RL} agent which, later, was extended under \ac{IRL} settings to include  implicit and explicit rewards, allowing for real-time adaptation from user social feedback. The approach was evaluated using a within-subjects \ac{HRI} study with $21$ participants with 
the results showing that both the explicit and implicit feedback mechanisms enabled an \textit{adaptive robo-waiter} that was rated as more \emph{enjoyable} and \emph{sociable} compared to the robot implementing the pre-trained model or using a random control policy. The adaptive robo-waiter also rendered more \textit{appropriate positioning} relative to the participants. Additionally, \textit{adaptability} ratings showed the explicit feedback condition as the most preferred condition with the robot being rated significantly higher in terms of \textit{understanding and adapting} to what the participant \textit{said}. These results clearly show that for task-based interactions (such as the robo-waiter context), adaptation based on affect alone is insufficient, and we do need to take into account explicit, task-related user feedback. Combining explicit and implicit feedback to shape the reward function, although not explored in this study, has the potential to further improve user interaction experience.

%

\subsection{Lesson \& Insights}

The creation of closed-loop affective robots that can undertake successful social interactions with humans requires that these robots keep learning in a lifelong manner and continually adapt towards user behaviours, their affective states and moods~\cite{churamani2020continual}. Traditional \ac{ML} approaches do not scale well to the dynamic nature of such real-world interactions because they often assume stationarity in data conditions and distributions, but real-world contexts change continuously. Also, the training data and learning objectives relevant to \ac{HRI} may change rapidly. The \acf{CL} paradigm is introduced in order to address these problems~\cite{thrun1995lifelong,parisi2019continual}. 

In~\cite{churamani2020continual}, we provide guidelines on how to utilise \ac{CL} for personalised affect perception and context-appropriate behavioural learning for affective robotics. These guidelines are then utilised in~\cite{Churamani-ACIIW2022}, to enable \ac{CL}-based \textit{personalisation} in the context of robotic wellbeing coaching, where a user study is conducted with $20$ participants comparing static and scripted interactions with using affect-based adaptation without personalisation, and using affect-based adaptation with \textit{continual personalisation}. The results showed that participants indicate a clear preference for a robotic coach with \textit{continual personalisation} capabilities, with significant improvements observed in the robot's \textit{anthropomorphism, animacy} and \textit{likeability} ratings. Additionally, the robot is also rated as significantly better at understanding how the participants felt during the interactions.

Although affective adaptation is a desirable capability for social robots, we need to bear in mind that it might not always work, and at times may even hinder interactions. For example, Kennedy~et~al.~\cite{KennedyEtAl-HRI15} investigated the effect of a social robot tutoring strategy, with and without social and adaptive behaviours, in the context of children learning about prime numbers. Their results showed \textit{no significant learning outcome} for children interacting with a robot using social and adaptive behaviours in addition to the teaching strategy. Therefore researchers should be cautious about the specific context they have at hand when deciding to apply social and adaptive behaviours to a robot, and whether these interactions are longitudinal or one-off. Gao~et~al.~\cite{GaoEtAl-ROMAN18} also investigated the effects of robot behaviour personalisation on user's task performance in the context of robot-supported learning. They utilised \ac{RL} for personalisation, enabling a robot tutor to select verbal supportive behaviours to maximise the user's task progress and positive reactions. Their results showed that participants were more efficient at solving logic puzzles and preferred a robot that exhibits more varied behaviours compared to a robot that personalises its behaviour by converging on a specific one over time. 
Overall, adaptation and personalisation based on affective or social behaviours needs further investigation, to gather insights on the impact of the context and the nature of the interaction (e.g., task-based \textit{vs.} free-flow). 



\begin{table}

\caption{Four \textit{Critical Lessons} with the \textit{Reflective Questions} the hitchhiker needs to probe when applying \acf{AC} solutions for \acf{HRI} research under situated interaction settings. }
\centering
\begin{tabular}{ | V| L| } 
\hline
\rowcolor{gray!40}
\makecell[c]{\textbf{Lesson}} & \makecell[c]{\textbf{Reflective Questions}}\\
\hline
    \rowcolor{gray!15}
    \textbf{Lesson 1:} The six basic emotion categories  are  irrelevant in situated interactions. &
    \vspace{-2mm}\begin{enumerate}[leftmargin=0.4cm]
        \itemsep-0.1em
        \item How is individual affective behaviour manifested?
        \item Which affect model can best represent an individual's affective states? 
    \vspace{-2mm} 
    \end{enumerate}
    
    \\
    \hline
    \rowcolor{gray!15}

    \textbf{Lesson 2:} Affect recognition accuracy (\%) improvements are unimportant. & 
    \vspace{-2mm}\begin{enumerate}[leftmargin=0.4cm]
        \itemsep-0.1em
        \item Are we placing the sensors in the right places? 
        \item Are we measuring the most relevant signals? 
        \item How can we interpret model performance in view of users' interaction experiences?
      \vspace{-2mm}
    \end{enumerate}
    \\
    \hline
    \rowcolor{gray!15}
    
    \textbf{Lesson 3:} Affect recognition does not generalise (well) across contexts. & 
    \vspace{-2mm}\begin{enumerate}[leftmargin=0.4cm]

        \itemsep-0.1em
        \item Are we taking into account contextual and person-specific aspects?
        \item Are we investigating for whom the trained models work well, and why? 
        \item How do model predictions work for specific user groups like children and elderly?
        \item Are there any strategies in place to mitigate prediction bias in models?
    \vspace{-2mm} 
    \end{enumerate}
    \\
    \hline
    \rowcolor{gray!15}

    \textbf{Lesson 4:}  Affect recognition alone is insufficient for adaptation and personalisation. & 
    \vspace{-2mm}\begin{enumerate}[leftmargin=0.4cm]
        \itemsep-0.1em
        \item Can the users' responses be summarised only using their affective behaviour?
        \item Is the user providing additional feedback that may be helpful for robot learning? 
        \item Is personalisation required and/or appropriate for the context of the interaction? 
    \vspace{-2mm} 
    \end{enumerate}
    \\
\hline
\end{tabular}
\label{table:lessons}
\vspace{-3mm}
\end{table}

\section{Summary and Conclusion}
\label{sec:conclusion}

This paper, reflecting upon the pitfalls and limitations of current \ac{AC} solutions for situated \ac{HRI}, presents \textit{four critical lessons} for the hitchhiker starting their \ac{HRI} research journey (see Table~\ref{table:lessons}). These lessons, learnt through our experience as well as a critical review of existing literature, aim to distill key challenges that need the focus and attention of the \ac{HRI} community. Specifically critiquing the use of \ac{AC} solutions for sensing, perceiving and understanding user affective behaviour in situated interactions, these lessons highlight the challenges and limitations of existing methodologies and how the hitchhiker needs to be cautious of utilising \textit{off-the-shelf} solutions, designed only for a generalised application that may not be efficient for situated interactions. For each lesson, we present a detailed account of the background and motivation for why it is relevant and provide contextual understanding of how it relates to pitfalls, in practice, of applying \textit{off-the-shelf} \ac{AC} solutions directly for situated \ac{HRI} studies. Furthermore, we also summarise our learning from these experiences and highlight key considerations for the hitchhiker to bare in mind. Our proposition, with this paper, is for the hitchhiker to reflect upon and probe specific questions, pertaining to each lesson, before designing \ac{HRI} studies that depend upon an affective evaluation of user behaviour. Furthermore, we also aim to encourage other researchers to contribute to the scientific community at large by critically analysing and reflecting upon \textit{`what does not work and why?'} in \ac{HRI} studies, and sharing their perspectives for everyone's benefit. Such a reflection and consideration may enable a successful and fruitful implementation of \ac{AC} solutions in situated interaction studies, \textit{creating a new bridge} for \ac{HRI}.



\bibliographystyle{IEEEtran}
\bibliography{HG-bib}

\begin{acronym}
    \acro{AC}{Affective Computing}
    \acro{AI}{Artificial Intelligence}
    \acro{AU}{Action Unit}
    \acro{AUC}{Area Under The Curve}
    \acro{AR}{Affective Robotics}
    \acro{CL}{Continual Learning}
    \acro{CNN}{Convolutional Neural Network}
    \acro{CV}{Computer Vision}
    \acro{EMG}{Electromyography}
    \acro{FACS}{Facial Action Coding System}
    \acro{FER}{Facial Expression Recognition}
    \acro{GAN}{Generative Adversarial Network}
    \acro{GEQ}{Game Experience Questionnaire}
    \acro{GEW}{Geneva Emotion Wheel}
    \acro{HCI}{Human-Computer Interaction}
    \acro{HRI}{Human-Robot Interaction}
    \acro{ICPR}{International Conference on Pattern Recognition}
    \acro{IRL}{Interactive Reinforcement Learning}
    \acro{LEAP-HRI}{Lifelong Learning and Personalization in Long-Term Human-Robot Interaction}
    \acro{ML}{Machine Learning}
    \acro{RL}{Reinforcement Learning}
    \acro{SAM}{Self-Assessment Manikin}
    \acro{SAR}{Socially Assistive Robotics}
    \acro{SER}{Speech Emotion Recognition}
    \acro{SOTA}{state-of-the-art}
    \acro{VR}{Virtual Reality}
    \acro{WM}{Working Memory}
    \acro{WAOW}{WorkingAge of Wellbeing}
\end{acronym}


\end{document}